\documentclass{article}

     \usepackage[nonatbib,preprint]{neurips_2020}
\usepackage[utf8]{inputenc}
\usepackage{cite}
\usepackage{amsmath,amssymb,amsfonts}
\usepackage{algorithmic}
\usepackage{graphicx}
\usepackage{textcomp}
\usepackage{times}  %Required
\usepackage{helvet}  %Required
\usepackage{courier}  %Required
\usepackage{url}  %Required
\usepackage{amsmath}
\usepackage{multirow}
\usepackage{paralist}
\usepackage{dsfont}
\usepackage{booktabs}
\usepackage{tabularx}
\input{taesup.sty}
\newcommand{\Pib}{\mathbf{\Pi}}
\newcommand{\Lb}{\mathbf{\Lambda}}
\newcommand{\Ellb}{\mathbf{L}}

\newcommand{\dude}{\texttt{2D-DUDE}}
\newcommand{\rand}{\texttt{2D-NDUDE(Rand)}}
\newcommand{\super}{\texttt{2D-NDUDE(Sup)}}
\newcommand{\supft}{\texttt{2D-NDUDE(Sup+FT)}}
\newcommand\ie{\textrm{i.e.}\ }
\newcommand\eg{\textit{e.g.}\ }

\usepackage{amsthm,amsmath}
\usepackage[utf8]{inputenc} %unicode support

\title{Supervised Neural Discrete Universal Denoiser for Adaptive Denoising}

\author{
Sungmin Cha\textsuperscript{\rm 1}\thanks{Equal contribution.},\ \ Seonwoo Min\textsuperscript{\rm 1}\footnotemark[1],\ \ Sungroh Yoon\textsuperscript{\rm 1},\ \ and Taesup Moon\textsuperscript{\rm 1} \\
  \textsuperscript{\rm 1} Department of Electrical and Computer Engineering, Seoul National University\\
  \texttt{\{sungmin.cha, mswzeus, sryoon, tsmoon\}@snu.ac.kr}
}

\begin{document}

\maketitle

\begin{abstract}
  We improve the recently developed Neural DUDE, a neural network-based adaptive discrete denoiser, by combining it with the supervised learning framework. Namely, we make the \emph{supervised pre-training} of Neural DUDE compatible with the \emph{adaptive fine-tuning} of the parameters based on the given noisy data subject to denoising. As a result, we achieve a significant denoising performance boost compared to the vanilla Neural DUDE, which only carries out the adaptive fine-tuning step with randomly initialized parameters. Moreover, we show the adaptive fine-tuning makes the algorithm robust such that a noise-mismatched or blindly trained supervised model can still achieve the performance of that of the matched model. Furthermore, we make a few algorithmic advancements to make Neural DUDE more scalable and deal with multi-dimensional data or data with larger alphabet size. We systematically show our improvements on two very diverse datasets, binary images and DNA sequences. 
\end{abstract}

\section{Introduction}
\label{sec:introduction}
\label{sec:intro}

The universal discrete denoising problem, which attempts to clean-up the noise-corrupted \emph{discrete} data without any assumptions on the clean source, was first proposed in \cite{Dude}. The paper devised Discrete Universal DEnoiser (DUDE) algorithm, which was rigorously shown to asymptotically achieve the performance of the best sliding window denoiser for \emph{all} possible sources, solely with the statistical knowledge on the noise mechanism. Such a powerful result of DUDE has led to many follow-up extensions in several areas \cite{Kamakshi,MotOrdRamSerWei11,OrdSerVerVis08,sdude,LeeMooYooWei16}. However, as pointed out in \cite{Buadasetal05}, the algorithm had some limitations that hindered it from being widely accepted in practice; the performance is sensitive to the sliding window size (\ie, context size) $k$ and tends to degrade as the alphabet size of the data increases.  

Recently, Neural DUDE \cite{MooMinLeeYoo16} has been proposed to remedy such limitations by combining the DUDE framework with deep neural networks. The main gist of \cite{MooMinLeeYoo16} is to define a single neural network model that can work on all possible contexts so that the information among similar contexts can be shared through the network parameters. Furthermore, the training of the neural network was done via devising ``pseudo-labels,'' which are computed solely based on the noisy observation data and the assumed noise model, and using them as target labels as in the multi-class classification. In results, the experiments in \cite{MooMinLeeYoo16} showed that Neural DUDE is much more robust to the context size $k$ and significantly outperforms DUDE for denoising binary images and DNA sequences. Also, \cite{park2020unsupervised} proposed Generalized CUDE (Gen-CUDE) which expands Neural DUDE from FIFO (Finite-input Finite-output) to FIGO (Finite-input General-output) and it proposed the theoretical bound which achieved the much tighter upper bound on the denoising performance compared with other baselines.

In this paper, we make further extensions of Neural DUDE to make it more effective, robust and scalable by making the following three contributions. First, we combine Neural DUDE with supervised learning and show that the supervised pre-training of Neural DUDE parameters on a separate training set could significantly improve its performance. Then, we show an adaptive fine-tuning of those pre-trained parameters with the given noisy data can be carried out using the pseudo-labels and achieve even better performance. In the later sections, we point out that such combination of supervised learning and adaptive fine-tuning is what differentiates Neural DUDE with a straightforward supervised learning based denoisers. 
% other methods; e.g., a straightforward supervised model that lacks adaptivity for the given noisy data or the recently developed CUDE which cannot combine with the supervised learning. 
Second, we show that the fine-tuning makes the algorithm robust such that a single blindly trained supervised model can adapt to various noise levels and data types without any performance loss. Such capability significantly can save the complexity for dealing with data with multiple noise levels. Third, we make additional algorithmic extensions beyond the original Neural DUDE such that it can handle multi-dimensional data as well as data with large alphabet size. 
% combining it with supervised learning. That is, we show that the supervised pre-training of Neural DUDE parameters on a separate training set could significantly improve its performance. Furthermore, we show that an adaptive fine-tuning of the pre-trained parameters with the given noisy data can be carried out using the pseudo-labels mentioned above. We show such fine-tuning combined supervised pre-training can not only further improve the performance of Neural DUDE, but also make the algorithm robust such that the potential mismatch of noise models between the supervised training set and the given noisy data can be corrected. Based on this robustness, we show a single blindly trained model can be developed 
% such that potential mismatch of noise models between the supervised trianing set and the given noisy data can be corrected to achieve 
% to the potential mismatch of noise models between the supervised training set and the given noisy data. 
We focus on systematically showing the effectiveness of our method with the experiments on the binary images and DNA sequences, and compare the improvements with the results given in \cite{MooMinLeeYoo16}.

% Such capability of adaptive fine-tuning is shown to be very effective, particularly when the mismatch of 

% the \emph{fine-tuning} of Neural DUDE with the given noisy data using the pseudo-labels can 

% so that the improvement over DUDE can be made even further by combining the original Neural DUDE with supervised training. We note that unlike the conventional supervised learning 

% Recently, a neural network-based method, dubbed as Neural DUDE \cite{MooMinLeeYoo16}, has been proposed and shown that the performance significantly surpasses that of previously proposed universal discrete denoiser, the Discrete Universal DEnoiser (DUDE) \cite{Dude}. The main gist of Neural DUDE is to devise \emph{pseudo-labels} solely based on the noisy data, hence train the neural network adaptively with given noisy data. In \cite{MooMinLeeYoo16}, the experimental results are given on two vastly different source of data, i.e., binary images and Oxford nanopore DNA sequences. 

% ,{}

%!TEX root = sup_n_dude.tex

\section{Preliminaries}

% Here, we introduce necessary notations and related work to precisely describe the algorithm in this paper. 
\subsection{Notations and Problem Setting}
We will mainly follow the notations in \cite{MooMinLeeYoo16}.
 % and refer the readers to the paper for full details.
 An $n$-tuple sequence is denoted as, \eg, $a^n=(a_1,\ldots,a_n)$, and $a_i^j$ refers to the subsequence $(a_i,\ldots,a_j)$. $\mathds{1}_a\in\mathbb{R}^n$ stands for the standard unit-vector in $\mathbb{R}^n$ that has 1 for coordinate $a$ and 0 for others. 
The upper and lower case letters stand for the random variables and either the realizations of the random variables or the individual symbols, respectively. 
% We denote $\Delta^d$ as the probability simplex in $\mathbb{R}^d$. 
The clean, underlying source data is denoted as an individual sequence $x^n$,
% as we do not assume any probabilistic models on it. 
and we assume each component $x_i$ takes a value in some finite set $\mcX$. For example, for binary data, $\mcX=\{0,1\}$, and for DNA data, $\mcX=\{\texttt{A},\texttt{C},\texttt{G},\texttt{T}\}$. 
% Furthermore, as usual, the uppercase letters stand for the random variables, and the lowercase letters stand for the realizations of the random variables or the individual (non-random) symbols. 
% We denote the underlying source data as $\{X_i\}$ and assume each component takes values in some finite set $\mcX$. 

We assume the noise mechanism is the Discrete Memoryless Channel (DMC), namely, the index-independent noise, and denote the noisy version of the source as $Z^n$, of which each $Z_i$ takes a value in, again, a finite set $\mcZ$.
% When the source sequence is corrupted by a Discrete Memoryless Channel (DMC), namely, the index-independent noise, it results in a noisy version of the source, $Z^n$, of which each $Z_i$ takes a value in, again, a finite set $\mcZ$.
% The resulting noisy version of the source corrupted by a DMC is denoted as $\{Z_i\}$, and its components take values in, again, some finite set $\mcZ$. 
The DMC is characterized by the channel transition matrix $\mathbf\Pi\in\mathbb{R}^{|\mcX|\times|\mcZ|}$, of which the $(x,z)$-th element stands for $\text{Pr}(Z=z|X=x)$.
% , \ie, the conditional probability of the noisy symbol taking value $z$ given the source symbol was $x$.
Following the settings in \cite{Dude,MooMinLeeYoo16}, we assume $\mathbf{\Pi}$ is of the \emph{full row rank} and \emph{known} to the denoiser. 

% An essential but natural assumption we make is that $\mathbf{\Pi}$ is of the \emph{full row rank}. 
% We also denote $\Pib^\dagger=\Pib^\top(\Pib\Pib^\top)^{-1}$ as the Moore-Penrose pseudoinverse of $\Pib$. 
% Furthermore, following the settings in \cite{Dude,MooMinLeeYoo16}, we assume $\Pib$ is \emph{known} to the denoiser. 

Given the noisy data $Z^n$, a discrete denoiser reconstructs the original data with $\hat{X}^n=\{\hat{X}_i(Z^n)\}_{i=1}^n$, where each $\hat{X}_i(Z^n)$ takes its value also in a finite set $\hat{\mathcal{X}}$. One representative form of such denoiser is the sliding-window denoiser, defined as $\hat{X}_i(Z^n)=s_k(Z_{i-k}^{i+k})$, in which $s_k:\mcZ^{2k+1}\rightarrow\hat{X}$ is a time-invariant mapping. We also denote the tuple $(Z_{i-k}^{i-1},Z_{i+1}^{i+k})\triangleq \Cb_i^k$ as the double-sided context around the noisy symbol $Z_i$. Moreover, we denote $\mcS\triangleq\{s:\mcZ\rightarrow\hat{\mcX}\}$ as the set of \emph{single-symbol denoisers}  that are sliding window denoisers with $k=0$. Note $|\mcS|=|\hat{\mcX}|^{|\mcZ|}$. Then, we can interpret $s_k(\Cb_i^k,\cdot)$ as a single-symbol denoiser in $\mcS$ determined by the context $\Cb_i^k$. The performance of $\hat{X}^n$ is measured by the average loss,
\[
L_{\hat{X}^n}(x^n,Z^n) = \frac{1}{n}\sum_{i=1}^n\Lb(x_i,\hat{X}_i(Z^n)),
% \label{eq:avg_loss}
\]
where $\Lb(i,j)$ is a loss function that measures the loss incurred by estimating $i\in\mcX$ with $j\in\mcXhat$. 
% The loss function is fully represented with a loss matrix $\mathbf{\Lambda}\in\mathbb{R}^{|\mcX|\times|\hat{\mcX}|}$. 

% \subsection{Related work}

\subsection{Neural DUDE}
\begin{figure}[h]
  \centering
  \includegraphics[width=0.6\textwidth]{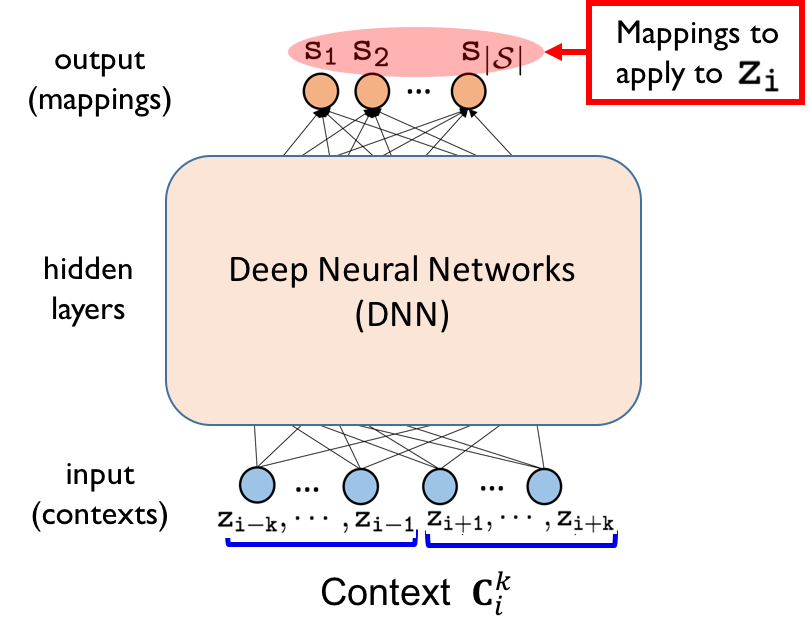}
%  \vspace{-.3cm}
  % \centerline{Example ar}\medskip
  \caption{An example architecture of $\mathbf{p}^k(\mathbf{w},\cdot)$.}
\label{fig:n_dude}%
\end{figure}

% DUDE \cite{Dude} is a universal denoiser based on context counting, and the specific denoising rule of DUDE is given in \cite[Eq.(11)-(13)]{Dude}. 
As mentioned in Introduction,
% , of which detailed denoising rule is given in \cite[Eq.(11)-(13)]{Dude}, 
a neural network-based sliding-window denoiser, Neural DUDE, has been recently proposed in \cite{MooMinLeeYoo16} to overcome the limitations of DUDE in \cite{Dude}. As exemplified in Figure \ref{fig:n_dude}, the algorithm defines a single fully connected neural network with parameters $\wb$,
\begin{eqnarray}
\mathbf{p}^k(\mathbf{w},\cdot):\mathcal{Z}^{2k}\rightarrow\Delta^{|\mcS|},\label{eq:neural_net}
\end{eqnarray} 
that works as a sliding-window denoiser. That is, at location $i$, the network takes the double-sided context $\Cb_i^k\in\mcZ^{2k}$ as input and outputs the probability distribution on the single-symbol denoisers in $\mcS$ to apply to $Z_i$. 
% \vspace{-.1in}

%  \vspace{-.3cm}

The reason for not including $Z_i$ as the input to the neural network is to derive ``pseudo-labels'' on the mappings $s\in\mcS$, based on the unbiased estimate of the expected true loss. That is, as detailed in \cite{MooMinLeeYoo16}, we can first devise an estimated loss
\begin{eqnarray}
\Ellb=\Pib^\dagger\bm{\rho}\in\mathbb{R}^{|\mcZ|\times|\mcS|},\label{eq:est_loss}
\end{eqnarray}
in which $\Pib^\dagger$ is the Moore-Penrose pseudoinverse of $\Pib$ and $\bm{\rho}\in\mathbb{R}^{|\mcX|\times|\mcS|}$ is a matrix such that its $(x,s)$-th element is 
\begin{eqnarray}
\bm\rho(x,s)=\mathbb{E}_{Z|x}\Lb(x,s(Z)). \label{eq:rho}
\end{eqnarray}
The notation $\mathbb{E}_{Z|x}(\cdot)$ stands for the expectation with respect to the distribution defined by the $x$-th row of $\Pib$, and we can show that $\Ellb$ has an unbiased property, $\mathbb{E}_{Z|x}\Ellb(Z,s)=\mathbb{E}_{Z|x}\Lb(x,s(Z))$. Then, Neural DUDE computes ``pseudo-label'' matrix $\Ellb_{\text{new}}\in\mathbb{R}^{|\mcZ|\times|\mcS|}$ as
\begin{eqnarray}
\Ellb_{\text{new}} \triangleq -\Ellb+L_{\text{max}}\mathbf{1}_{|\mcZ|}\mathbf{1}_{|\mcS|}^\top,\label{eq:L_max}
\end{eqnarray}
in which $L_{\text{max}}\triangleq \max_{z,s}\Ellb(z,s)$, and $\mathbf{1}_{|\mcZ|}$ and $\mathbf{1}_{|\mcS|}$ stand for the all-1 vector with $|\mcZ|$ and $|\mcS|$ dimensions, respectively. 
% Note $\Ellb_{\text{new}}$ can be computed solely based on $\Lb$ and the assumption of known $\mathbf{\Pi}$. 
Note that all the elements in $\Ellb_{\text{new}}$ can be computed with $z$ and $s$ (and \emph{not} with $x$) and are designed to be non-negative. 

Now, in order to train $\wb$, Neural DUDE then uses $\mathbf{L}_{\text{new}}^\top\mathds{1}_{Z_i}\in\mathbb{R}_+^{|\mcS|}$ as a pseudo-label vector, which is no longer a one-hot vector, for the single-symbol denoiser to apply to $Z_i$ that has context $\Cb_i^k$.  The parameters are learned by minimizing the objective function in \cite[Eq.(7)]{MooMinLeeYoo16},
\begin{eqnarray}
\mathcal{L}(\mathbf{w},Z^n)&\triangleq&\frac{1}{n}\sum_{i=1}^n\mathcal{C}\Big(\mathbf{L}_{\text{new}}^\top\mathds{1}_{Z_i},\mathbf{p}^k(\mathbf{w},\mathbf{C}_i^k)\Big),\label{eq:loss_function_for_nn}
\end{eqnarray}
in which $
\mathcal{C}(\mathbf{g},\mathbf{p})\triangleq -\sum_{i=1}^{|\mcS|} g_i\log p_i\label{eq:alter_obj_fcn}$ for $\mathbf{g}\in\mathbb{R}_+^{|\mcS|}$ and $\mathbf{p}\in\Delta^{|\mcS|}$. 
% For minimizing (\ref{eq:loss_function_for_nn}), the ordinary optimization methods, \eg, back-propagation and the variants of mini-batch stochastic gradient descent (SGD) \cite{Zei12,TieHin12}, are used.  
Once (\ref{eq:loss_function_for_nn}) converges after sufficient number of mini-batch SGD based updates, \eg, \cite{Zei12,TieHin12,KinBa15}, the converged parameter $\wb^*$ is used to obtain the resulting sliding window denoiser $s_{k,\text{Neural DUDE}}(\Cb^k,\cdot)=\arg\max_s\mathbf{p}(\wb^*,\Cb^k)$ for $\Cb^k\in\mcZ^{2k}$. The reconstruction by Neural DUDE at location $i$ is then defined by 
\[
\hat{X}_{i,\text{Neural DUDE}}(Z^n)=s_{k,\text{Neural DUDE}}(\Cb^k_i,Z_i).
\]
We note Neural DUDE is \emph{adaptive} since it trains the neural network parameters solely with the given noisy data $Z^n$ subject to denoising. 
% the above training process is \emph{adaptive} with respect to the noisy observation as computing (\ref{eq:loss_function_for_nn}) does not require any additional data other than $Z^n$.
For the complete details on Neural DUDE, we refer the readers to \cite{MooMinLeeYoo16}.

% \subsubsection{CUDE}

% The effectiveness of Neural DUDE has been shown via the experiments in \cite{MooMinLeeYoo16} on binary image denoising and DNA sequence denoising. On both applications, Neural DUDE was shown to significantly outperform DUDE, which has already been shown to achieve the state-of-the-art performance, \eg, \cite{Ordetal03,LeeMooYooWei16,MoWeKi11}. In this paper, we further improve Neural DUDE via supervised learning and show that our improvement gives another significant performance boost. 

% while DUDE was already achieving the state-of-the-art performance \cite{Ordetal03,LeeMooYooWei16,MoWeKi11}, and Neural DUDE significantly outperforms DUDE. 

% Tha algorithm devised ``pseudo-labels'' $\Ellb_{\text{new}}$, which can be solely computed with noisy data and $\Pib$, and use them as target labels for learning the network parameters, as shown in the objective function \cite[Eq.(7)]{MooMinLeeYoo16}. In \cite{MooMinLeeYoo16}, the randomly initialized weights are used, and in this paper, we try to initialize with supervised pre-training. 

%!TEX root = sup_n_dude.tex

\section{Methods}

\subsection{Supervised pre-training and adaptive fine-tuning}

While the universal setting in \cite{Dude} assumes that the underlying clean source can be any arbitrary sequence, many practical scenarios target developing denoisers tailored for the given applications. Hence, it is natural to apply supervised learning (\eg, \cite{BurSchHar12}) for training a denoiser before applying it to the given noisy data, by collecting additional supervised dataset from the considered application. Collecting such dataset would not be difficult since we assume to know $\Pib$; \ie, we can collect representative clean data $\tilde{x}^n$ from the considered application, \eg, benchmark images or reference DNA sequences from several species, and corrupt them with the assumed $\Pib$ to generate the noisy counterpart $\tilde{Z}^n$. Once we generate such clean-noisy data pairs, we can then generate a supervised training data,
% the (clean symbol,noisy context) pairs,
\[
\mathcal{D}\triangleq\{(\tilde{x}_i,(\tilde{\Cb}_{i}^k,\tilde{Z}_i))\}_{i=1}^N,\label{eq:supervised_set}
\]
 in which $\tilde{\Cb}^k_i\in\mcZ^{2k}$ is for the $k$-th order double-sided context around the noisy symbol $\tilde{Z}_i$, $\tilde{x}_i$ is the clean underlying symbol corresponding to $\tilde{Z}_i$, and $N$ is the total number of such collected pairs\footnote{For notational clarity, we used the \emph{tilde} notation for the supervised data to differentiate them from the given noisy data $Z^n$ subject to denoising.}.

Now, the subtle point for training a neural network-based sliding-window denoiser with $\mathcal{D}$ is that, instead of directly learning a mapping $\mcZ^{2k+1}\rightarrow\mcXhat$, we remain in using the network form of Neural DUDE, (\ref{eq:neural_net}). That is, we first define $\bm\rho_{\text{true}}
\in\mathbb{R}^{|\mcX||\mcZ|\times|\mcS|}$ such that 
\[
\bm\rho_{\text{true}}((x,z),s) = \Lb(x,s(z)),
\]
in which $(x,z)\in\mcX\times\mcZ$ and $s\in\mcS$ index the rows and columns of $\bm\rho_{\text{true}}$, respectively. Then, we define 
\[\Ellb_{\text{true}} = -\bm\rho_{\text{true}} + \Lambda_{\max}\mathbf{1}\mathbf{1}^\top\] where $\Lambda_{\max}=\max_{x,\hat{x}}\Lb(x,\hat{x})$. Using $\Ellb_{\text{true}}^\top\mathds{1}_{(\tilde{x}_i,\tilde{Z}_i)}\in\mathbb{R}^{|\mcS|}_+$ as a ground-truth label for the single-symbol mapping at location $i$, we can then learn $\wb$ by minimizing 
% the following objective function for the supervised training, 
\begin{eqnarray}
\mathcal{L}_{\text{supervised}}(\wb, \mathcal{D})\triangleq \frac{1}{N}\sum_{i=1}^N \mathcal{C}\Big(\Ellb_{\text{true}}^\top\mathds{1}_{(\tilde{x}_i,\tilde{Z}_i)},\mathbf{p}^k(\mathbf{w},\tilde{\Cb}_i^k) \Big). \label{eq:supervised_loss}
\end{eqnarray}
Note $\Ellb_{\text{true}}^\top\mathds{1}_{(\tilde{x}_i,\tilde{Z}_i)}$ may not be a one-hot vector as in the standard supervised classification, because the loss function $\Lb$ can be arbitrary and multiple single-symbol denoisers can achieve the minimum loss value for the given pair $(\tilde{x}_i,\tilde{Z}_i)$. 
% Nonetheless, we can see that minimizing (\ref{eq:supervised_loss}) would result in learning parameters that will minimize the average true loss on $\mathcal{D}$ when used for denoising as in Neural DUDE. 
% Again, minimizing (\ref{eq:supervised_loss}) is done by the standard back-propagation and variants of the mini-batch SGD. 

% (\ref{eq:supervised_loss}) looks identical to the objective function of the standard supervised multi-class classification for neural networks with input-output pairs $(\tilde{\Cb}_i^k,\tilde{s}_i)$. Again, minimizing (\ref{eq:supervised_loss}) is done by the standard back-propagation and variants of the mini-batch SGD. 

The reason for learning the neural network of the form (\ref{eq:neural_net}) during the supervised training is because it becomes compatible with the adaptive fine-tuning step using the given noisy data $Z^n$ subject to denoising. That is, by denoting $\tilde{\wb}$ as the converged parameter after minimizing (\ref{eq:supervised_loss}), we can further adaptively update (\ie, fine-tune) $\tilde{\wb}$ tailored for $Z^n$ by minimizing $\mathcal{L}(\wb,Z^n)$ in (\ref{eq:loss_function_for_nn}). Note $\tilde{\wb}$ can be interpreted as a pre-trained initialization of Neural DUDE in contrast to the random initialization used in the original paper \cite{MooMinLeeYoo16}. 
% We denote the final denoiser obtained after the fine-tuning as \texttt{NDUDE(Sup(Blind)$+$FT)}.
% once the objective function  (\ref{eq:supervised_loss}) converges after sufficient iterations of weight updates, we denote the converged parameter as $\tilde{\wb}$. Then, for a given noisy data to denoise, $Z^n$, we can further update $\tilde{\wb}$ adaptively for $Z^n$ by minimzing $\mathcal{L}(\wb,Z^n)$ in (\ref{eq:loss_function_for_nn}) starting from $\tilde{\wb}$, hence, fine-tuning. 

As mentioned in the Introduction, we show in the experimental section that the framework of combining supervised pre-training with adaptive fine-tuning, denoted as  \texttt{NDUDE(Sup(Blind)$+$FT)}, becomes very effective in improving the denoising performance as well as achieving the robustness of the algorithm. We carry out systematic ablation studies and also compare our method with a \emph{vanilla} supervised-only model, which directly learns a mapping $\mcZ^{2k+1}\rightarrow\mcXhat$ and possesses no adaptivity, and show the superiority of \texttt{NDUDE(Sup(Blind)$+$FT)}. 

% The effectiveness of the combination of pre-training with fine-tuning would be clearly shown in the experimental section. 

% That is, we can collect representative clean data $\tilde{x}^n$ from the given application, \eg, benchmark images or standard DNA sequences from several species, and corrupt them with the assumed $\Pib$ to generate the noisy counterpart $\tilde{Z}^n$. For training sliding-window denoisers, we can generat

% \be
% \mathcal{D}\triangleq\{(\tilde{x}_i,\tilde{\Cb}_{i})\}_{i=1}^m
% \ee, 

% The main contribution of \cite{MooMinLeeYoo16} was to devise pseudo-labels such that the neural network can be trained with the very noisy data subject to denoising. 

\subsection{2D contexts and output dimension reduction}

% In this section, we give several algorithmic improvements specifically tailored for images and DNA sequences. 

In \cite{MooMinLeeYoo16}, all the experiments only dealt with one dimensional (1D) data, including the raster scanned images. However, for multi-dimensional data like images, it is clear to directly use the multi-dimensional context. Hence, in our experiments with images, we used two-dimensional (2D) contexts as in \cite{MoWeKi11,OrdSerVerWeiWei03}. 
% For images, instead of the one dimensional (1D) context from raster scanning as used in \cite{MooMinLeeYoo16}, we used two-dimensional (2D) contexts as in \cite{MoWeKi11,OrdSerVerWeiWei03}, which are more natural for images.
% would work to some extent, it is more natural and effective to use two-dimensional (2D) contexts \cite{MoWeKi11,OrdSerVerWeiWei03}.
% Such extension for the original DUDE to two-dimensional data using 2D contexts is given in \cite[Fig. 1]{MoWeKi11} by defining a pre-ordered 2D context sequence around the given data. 
To that end, we use $\Cb_i^{\ell\times \ell}$, an $\ell\times\ell$ square patch around the noisy symbol $Z_i$ that does \emph{not} include $Z_i$, as the 2D context. 
% the square-patch \emph{with a hole} as the 2D context; \ie, we define $\Cb_i^{\ell\times \ell}$ as the $\ell\times\ell$ patch around the noisy symbol $Z_i$ that does \emph{not} include $Z_i$. 
Note when odd values for $\ell$ are used, the 1D context $\Cb_i^k$ with $k=(\ell^2-1)/2$ has the same data size as $\Cb_i^{\ell\times \ell}$.
% Thus, when we use odd values for $\ell$, $\Cb_i^{\ell\times \ell}$ is equivalent to the 1-D context with $k=(\ell^2-1)/2$ in terms of the input data size to the neural network. 
Furthermore, we did the standard zero-padding of size $(\ell-1)/2$ at the boundary of the image to utilize the partial 2D contexts at the boundaries. 
\begin{figure}[htb]

\begin{minipage}[b]{1\linewidth}\label{fig:original}
  \centering
  \centerline{\includegraphics[width=0.8\linewidth]{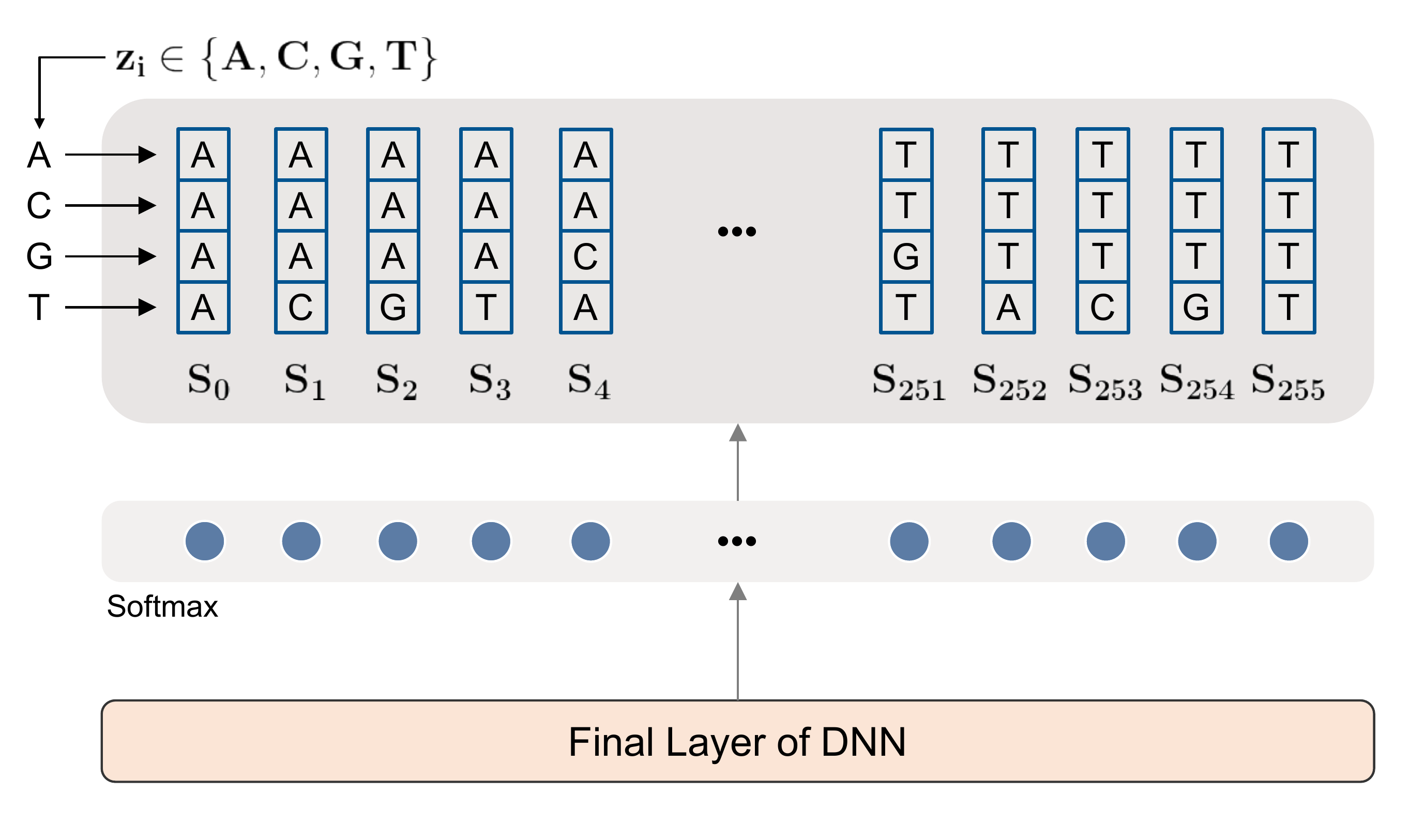}}
%  \vspace{2.0cm}
  \centerline{(a) Original output layer}\medskip
\end{minipage}\vspace{-.5in}
\begin{minipage}[b]{1\linewidth}
  \centering
  \centerline{\includegraphics[width=0.8\linewidth]{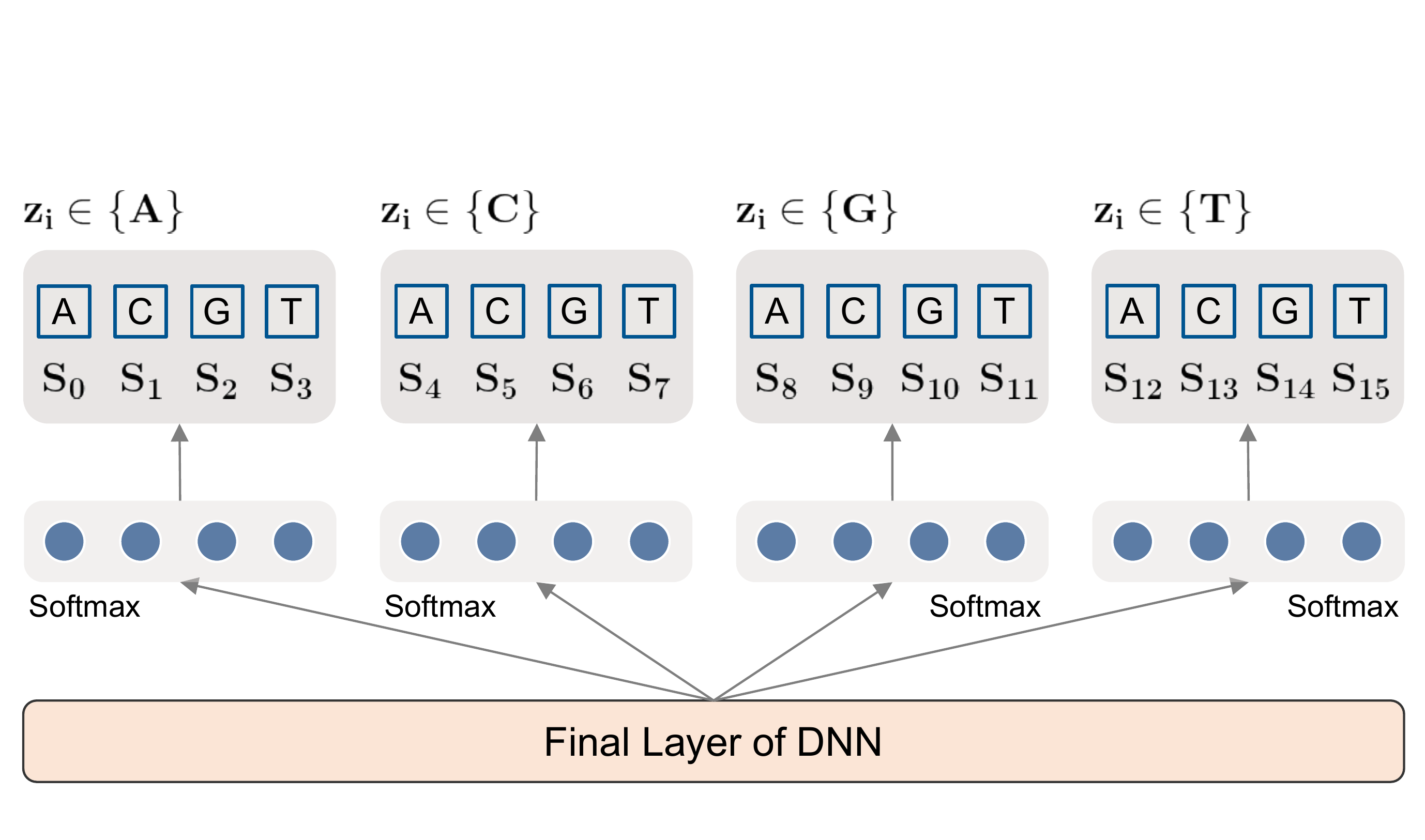}}
%  \vspace{1.5cm}
  \centerline{(b) Reduced output layer}\medskip
\end{minipage}\vspace{-.15in}
\caption{Output reduced architecture for DNA data.}
\label{fig:reduced_arch}
\end{figure}

Moreover, one caveat of the original Neural DUDE is that the output size of the network $\mathbf{p}(\wb,\cdot)$ can exponentially increase when the alphabet size of the data increases. For example, even for DNA sequence that has alphabet size of 4, the output size of $\mathbf{p}(\wb,\cdot)$ becomes $|\mcS|=|\hat{\mcX}|^{|\mcZ|}=256$. Hence, to make Neural DUDE more scalable for larger alphabet data, it is necessary to reduce the output dimension of Neural DUDE. Here, we show we can significantly reduce the output dimension to $|\mcZ|\cdot|\hat{\mcX}|$ instead of $|\hat{\mcX}|^{|\mcZ|}$.
% For DNA sequences, we reduced the output size of the network $\mathbf{p}(\wb,\cdot)$ in (\ref{eq:neural_net}) from $|\mcS|=|\hat{\mcX}|^{|\mcZ|}=256$ to $|\mcZ|\times|\hat{\mcX}|=16$ for a more efficient learning of the network. Note $|\mcZ|=|\mcX|=|\hat{\mcX}|=4$ for DNA sequence. 
% While learning the model with 256 output units is doable, we reduce the output size to more efficiently learn the neural network model. 
% to reduce the number of parameters to learn. Since $|\mcZ|=|\mcX|=|\hat{\mcX}|=4$, the vanilla Neural DUDE has to define a network with output size $|\mcS|=|\hat{\mcX}|^{|\mcZ|}=256$. While such output size could be still learnable, we instead reduce the output size to $|\mcZ|\times|\hat{\mcX}|=16$ to more efficiently learn the denoiser. 

The main observation for such reduction is to decompose a single-symbol denoiser $s\in\mcS$ to mappings for each noisy symbol $z\in\mcZ$. For examples, for the case of DNA data, if we specify $\hat{x}\in\{\texttt{A},\texttt{C},\texttt{G},\texttt{T}\}$ for each $z\in\{\texttt{A},\texttt{C},\texttt{G},\texttt{T}\}$, then we can completely enumerate all the single-symbol denoisers in $\mcS$. Figure \ref{fig:reduced_arch} compares the reduced output layer with the original architecture for DNA data. 

% (b) shows our reduced output layer architecture that has 4 outputs for each $z\in\{\texttt{A},\texttt{C},\texttt{G},\texttt{T}\}$ symbol as opposed to Figure \ref{fig:reduced_arch}(b) that has 256 outputs. 
% To realize such specification, instead of defining 4 different models, one for each $z\in\{\texttt{A},\texttt{C},\texttt{G},\texttt{T}\}$ symbol, we define a single model that has 4 output layers (that share the lower layers), as shown in Figure \ref{fig:reduced_arch}(b).
% \vspace{-.15in}

% \vspace{-.1in}

% As shown in Figure \ref{fig:reduced_arch}(b), the reduced network architecture would thus have 4 output layers (that share the lower layers) corresponding to each $z\in\{\texttt{A},\texttt{C},\texttt{G},\texttt{T}\}$ symbol. 
Now, for the supervised pre-training of the reduced network, given a data point $(\tilde{x}_i,(\tilde{\Cb}_i^k,\tilde{Z}_i))$ in $\mathcal{D}$, we 
% first compute 
% $$\tilde{\hat{x}}_i=\arg\min_{\hat{x}=s(\tilde{Z})}\Lb(\tilde{x}_i,s(\tilde{Z}_i))
% $$ 
use $\mathds{1}_{\tilde{x}_i}\in\mathbb{R}^{|\mcXhat|}$ as the target label for the output layer corresponding to $\tilde{Z}_i$. For other parts of the output layers that do not correspond to $\tilde{Z}_i$, we use the all-1 vector, $\mathbf{1}\in\mathbb{R}^{|\mcXhat|}$
, as target label such that uniform output across $\mcXhat$ is encouraged. 
% For supervised training of the network, given a data point $(\tilde{x}_i,(\tilde{\Cb}_i^k,\tilde{Z}_i))$, we obtain 

For the adaptive fine-tuning, we first compute a matrix $\bm\rho_z\in\mathbb{R}^{|\mcX|\times|\mcXhat|}$ such that 
\begin{eqnarray}
\bm\rho_z(x,\hat{x})=\Pib(x,z)\Lb(x,\hat{x}), \label{eq:partial}
\end{eqnarray}
for each $z\in\mcZ$. By denoting $\hat{x}=s(z)$ in (\ref{eq:partial}), we can see that $\bm\rho(x,s)$ defined in (\ref{eq:rho}) can be expressed as
\begin{eqnarray}
\bm\rho(x,s) = \sum_{z}\bm\rho_z(x,\hat{x}).\label{eq:partial_rho}
\end{eqnarray}
Now, we define $
\Ellb_{z} = \Pib^{\dagger}\bm\rho_{z}\in\mathbb{R}^{|\mcZ|\times |\mcXhat|}
$ and \[\Ellb_{\text{new},z}=-\Ellb_{z} + L_{\max}\mathbf{1}\mathbf{1}^\top\] with $L_{\max}=\max_{z,z',\hat{x}}\Ellb_z(z',\hat{x})$. From (\ref{eq:partial_rho}), we can see $\Ellb$ in (\ref{eq:est_loss}) can be expressed as $\sum_z\Ellb_z$, hence, $\Ellb_{\text{new},z}$ can be interpreted as a ``partial'' pseudo-label matrix associated with the noisy symbol $z$. Now, given a pair $(\Cb_i^k,Z_i)$, the pseudo-label for the output module corresponding to each $z\in\mcZ$ is then given as $\Ellb_{\text{new},z}^\top\mathds{1}_{Z_i}$. Note this pseudo-label is also solely defined from the noisy observation and $\Pib$, hence, we can see that the adaptive fine-tuning becomes possible for the reduced network.

\section{Results and Discussion}

\subsection{Binary image desnoising}\label{sec:image_denoising}
% We repeat the experiments in \cite[Section 5.2]{MooMinLeeYoo16} and report the relative improvements compared to the original Neural DUDE. 
% Furthermore, we show the effectiveness of the adaptive fine-tuning by showing the robustness to the potential mismatch in the assumed noise model in the supervised training set. 
% not only the improvement over the supervised learning-only, but also the robustness to the mismatch in the assumed noise model in the supervised training set. 

% ICASSP 2018 제출내용
% We used the 5 binary images identical to those used in \cite[Section 5.2]{MooMinLeeYoo16}
% % , \ie, halftone Einstein, Lena, barbara, Cameramen, and scanned Shannon paper images, 
% as test images for denoising to directly compare the results with the original Neural DUDE. Moreover, 8 other benchmark images were used as a validation set to tune the hyperparameters, \eg, learning rate, number of epochs, and the context size $k$, etc. For the supervised training dataset, $\mathcal{D}$, we used binarized 300 natural images from the Berkeley Segmentation Dataset \cite{berkeley}. $\Pib$ was the binary symmetric channel with $\delta$, and $\Lb$ was the Hamming loss.

% 차성민 작성

\subsubsection{Datasets and training details}
We first carry out experiments on denoising binary images that are corrupted by binary symmetric channel (BSC) with crossover probability $\delta$. We experimented with 4 different noise levels, $\delta=\{0.05, 0.1, 0.2, 0.25\}$. As in \cite{MooMinLeeYoo16}, we used Hamming loss as $\Lb$, thus, the average loss becomes Bit Error Rate (BER). For generating the supervised training dataset, $\mathcal{D}$, we took 50 binarized natural images from the training dataset of the Berkeley Segmentation Dataset (BSD) \cite{berkeley} as clean images. For testing the denoising performance, we used three separate datasets, Set11, Set2, and BSD20. Set11 consists of broadly used benchmark images in the image processing area, e.g., Barbara, Lena, and C.man, etc. In contrast, Set2 consists of two images, Einstein (a halftone image) and Shannon (a scanned text image), of which characteristics are radically different from those in Set11 and from the training dataset. We also selected 20 test images from the test dataset of the BSD, and we named it as BSD20.

All experiments were done with using Keras with Tensorflow\footnote{\texttt{http://keras.io, http://www.tensorflow.org}} backend. We used fully-connected neural network that had 12 layers and 128 nodes for each layer for our Neural DUDE, and the usual Rectified Linear Unit (ReLU) was used as the activation function. He initialization \cite{prelu} was used for initializing the weight parameters, and we did not use the weight decay regularization. 
For optimizer, Adam \cite{KinBa15} with learning rate of $10^{-3}$ was used for both supervised training and fine-tuning, and we did not use any learning rate decay. We denote our Neural DUDE as \texttt{NDUDE} for brevity. 

In order to validate the effect of the supervised training, we trained following three types of supervised models.
\begin{compactitem}
\item Noise-specific (\texttt{Sup}): These are models dedicated to each noise level, obtained by training on the separate training set associated with each $\delta$. 
\item Blindly-trained (\texttt{Sup(Blind)}): This is a \textit{single} model trained from composite noise levels; at the beginning of each training iteration, noisy images using a randomly selected $\delta$ in the range of $[0.05,0.25]$ are generated, and the model is trained with these images.
\item Vanilla-supervised (\texttt{SL}): These are also the models dedicated to each noise level, but they learn direct mappings, $\mcZ^{2k+1}\rightarrow \mcXhat$. Thus, they cannot adapt to the given noisy image via fine-tuning as in Neural DUDE.  
\end{compactitem}
In results, we compare total 9 models: two 1D models and seven 2D models. 
 % that are compared in the figure. 
% In the figure, we compared 9 schemes in total, which we describe below.
As baselines, we first reproduced \texttt{1D-DUDE} and \texttt{1D-NDUDE(Rand)} from \cite{Dude} and \cite{MooMinLeeYoo16}, respectively. The \texttt{Rand} notation stands for the scheme that trains with randomly initialized weight parameters. For 2D models, \texttt{2D-DUDE} and \texttt{2D-NDUDE(Rand)} are implemented as their 1D counterparts. Moreover, for the supervised-only models, \super, \texttt{2D-NDUDE (Sup(Blind))}, and \texttt{2D-SL} are compared. Finally, the two fine-tuned Neural DUDE models, \supft and \texttt{2D-NDUDE (Sup(Blind)$+$FT)}, are compared to test the adaptivity with respect to the given noisy images and the efficiency of the blindly trained model.

\begin{table*}[h]
\caption{Detailed denoising results on four images in Set11. (\textbf{Boldface} denotes the best BER in each row.)}
\centering
\smallskip\noindent
\resizebox{0.97\linewidth}{!}{
\begin{tabular}{|c|c||c|c|c|c|c|c|c|c|c|}
\hline
Error rate / $\delta$            & Image   & 1D-DUDE & 1D-NDUDE & 2D-DUDE & \begin{tabular}[c]{@{}c@{}}2D-NDUDE\\ (Rand)\end{tabular} & \begin{tabular}[c]{@{}c@{}}2D-NDUDE\\ (Sup)\end{tabular} & \begin{tabular}[c]{@{}c@{}}2D-NDUDE\\ (Sup+FT)\end{tabular} & \begin{tabular}[c]{@{}c@{}}2D-NDUDE\\ (Sup(Blind))\end{tabular} & \begin{tabular}[c]{@{}c@{}}2D-NDUDE\\ (Sup(Blind)+FT)\end{tabular} & 2D-SL    \\ \hline
\hline
\multirow{4}{*}{$\delta$ = 0.05} & C.man   & 0.507   & 0.462    & 0.426   & 0.350                                                     & 0.322                                                    & 0.331                                                        & 0.657                                                           & \textbf{0.321}                                                              & 0.330 \\ 
                                 & Barbara & 0.519   & 0.478    & 0.360   & 0.252                                                     & 0.296                                                    & 0.257                                                        & 0.599                                                           & \textbf{0.248}                                                             & 0.302 \\ 
                                 & F.print & 0.579   & 0.564    & 0.404   & 0.373                                                     & 0.382                                                   & \textbf{0.363}                                                        & 0.424                                                           & 0.365                                                              & 0.400 \\ 
                                 & Lena    & 0.320   & 0.296    & 0.180   & 0.158                                                     & 0.150                                                    & 0.154                                                        & 0.183                                                           & \textbf{0.148}                                                              & 0.156 \\ \hline
                                 \hline
\multirow{4}{*}{$\delta$ = 0.10} & C.man   & 0.484   & 0.442    & 0.409   & 0.329                                                     & 0.293                                                    & 0.290                                                        & 0.352                                                           & \textbf{0.286}                                                              & 0.301 \\ 
                                 & Barbara & 0.493   & 0.450    & 0.360   & 0.235                                                     & 0.279                                                    & 0.237                                                        & 0.363                                                           & \textbf{0.230}                                                              & 0.282 \\ 
                                 & F.print & 0.573   & 0.551    & 0.380   & 0.354                                                     & 0.368                                                    & \textbf{0.345}                                                        & 0.369                                                           & 0.347                                                              & 0.376 \\ 
                                 & Lena    & 0.297   & 0.277    & 0.164   & 0.142                                                     & \textbf{0.128}                                                    & 0.129                                                        & 0.139                                                           & \textbf{0.128}                                                              & 0.134 \\ \hline
                                 \hline
\multirow{4}{*}{$\delta$ = 0.20} & C.man   & 0.481   & 0.395    & 0.440   & 0.311                                                     & 0.275                                                    & 0.275                                                        & 0.282                                                           & \textbf{0.273}                                                              & 0.279 \\ 
                                 & Barbara & 0.507   & 0.455    & 0.394   & 0.253                                                     & 0.275                                                    & 0.248                                                        & 0.277                                                           & \textbf{0.234}                                                              & 0.276 \\ 
                                 & F.print & 0.642   & 0.603    & 0.448   & 0.379                                                     & 0.389                                                    & 0.363                                                        & 0.413                                                           & \textbf{0.355}                                                              & 0.401 \\ 
                                 & Lena    & 0.334   & 0.292    & 0.209   & 0.134                                                     & 0.125                                                    & 0.124                                                        & 0.129                                                           & \textbf{0.122}                                                              & 0.126 \\ \hline
                                 \hline
\multirow{4}{*}{$\delta$ = 0.25} & C.man   & 0.532   & 0.404    & 0.485   & 0.315                                                     & \textbf{0.278}                                                    & 0.283                                                        & 0.299                                                           & 0.283                                                              & 0.281 \\ 
                                 & Barbara & 0.538   & 0.457    & 0.440   & 0.263                                                     & 0.271                                                    & 0.248                                                        & 0.290                                                           & \textbf{0.238}                                                              & 0.270 \\ 
                                 & F.print & 0.689   & 0.513    & 0.638   & 0.400                                                     & 0.423                                                    & 0.384                                                        & 0.462                                                           & \textbf{0.373}                                                              & 0.432 \\
                                 & Lena    & 0.394   & 0.274    & 0.323   & 0.138                                                     & 0.127                                                    & \textbf{0.125}                                                        & 0.158                                                           & \textbf{0.125}                                                              & 0.128 \\ \hline
\end{tabular}}
\label{table1}
\end{table*}\vspace{-.25cm}

\subsubsection{Results}

\begin{figure*}[h]
\centering
\includegraphics[width=0.8\textwidth]{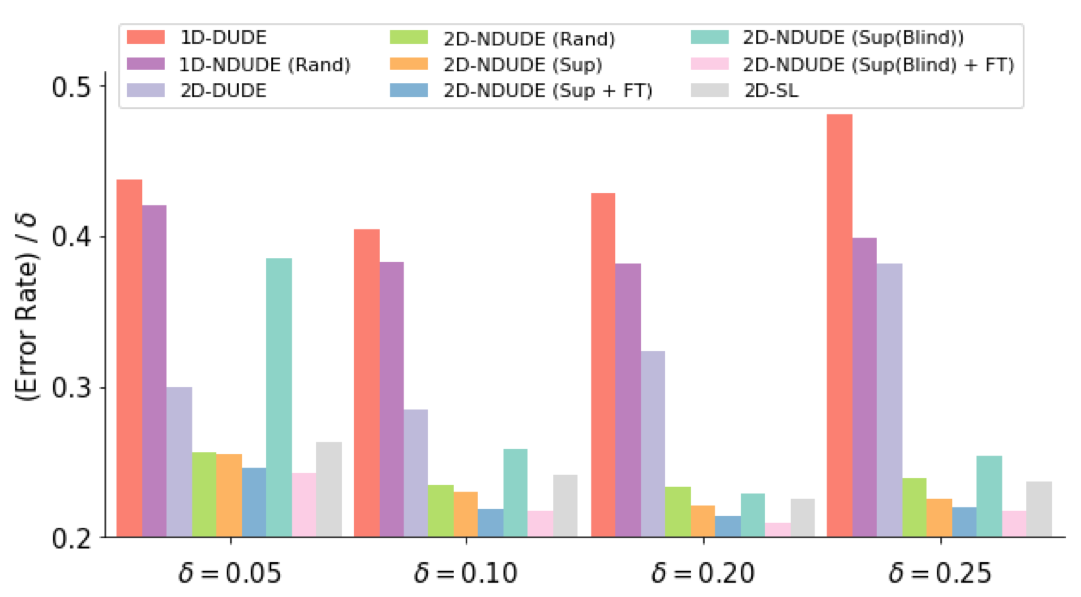}
\caption{BER results on Set11 with several $\delta$'s.}\label{fig:set11}
\end{figure*}

Figure \ref{fig:set11} shows the average denoising results on the Set11 benchmark for 4 different noise levels. The vertical axis is the Bit Error Rate (BER) normalized with $\delta$ after denoising, and all results are for the best $k$ values for each scheme.

% , we give denoising results on Set11 for 4 different noise levels, $\delta$=$\{0.05, 0.1, 0.2, 0.25\}$. 
% Figure \ref{fig:set11} shows the denoising results on the Set11. 

From the figure, we make the following observations. Firstly, by comparing 1D and 2D methods, we clearly see that the 2D contexts are much more effective than 1D contexts for denoising images. Also, paralleling the results of \cite{MooMinLeeYoo16}, we observe \rand  \ again significantly outperforms \dude. Second, we observe that supervised training is indeed useful for Neural DUDE as \texttt{2D-NDUDE(Sup)} outperforms \texttt{2D-NDUDE(Rand)}. The vanilla \texttt{2D-SL} also is quite strong as it sometimes outperforms \texttt{2D-NDUDE(Rand)}, although it is always slightly worse than \texttt{2D-NDUDE (Sup)}. Third, we confirm the effectiveness of the adaptive fine-tuning as \texttt{2D-NDUDE(Sup$+$FT)} always improves \texttt{2D-NDUDE(Sup)} on every $\delta$. This shows the additional adaptiveness is effective even for the matched supervised models. Finally,  \texttt{2D-NDUDE(Sup(Blind)$+$FT)}, which fine-tunes a base supervised model \texttt{2D-NDUDE(Sup(Blind))} with correct $\delta$, achieves impressive denoising results and even slighlty outperforms \texttt{2D-NDUDE(Sup$+$FT)}, which has matched supervised models for each noise level. 
Note this result is practically meaningful as it suggests to just maintain a \emph{single} supervised model as long as one can utilize the true noise level $\delta$ during the fine-tuning stage. 
Table \ref{table1} shows the actual BER values on the four popular images in Set11. We clearly observe that on most cases and $\delta$'s, \texttt{2D-NDUDE(Sup(Blind)$+$FT)} achieves the best BER results, whereas \texttt{2D-NDUDE(Sup(Blind))} sometimes suffers from high error rates. Moreover, depending on the images, the fine-tuned models significantly outperforms the supervised-only models,  \texttt{2D-NDUDE(Sup)} or \texttt{2D-SL}.

\begin{figure*}[t!]
\centering
  \includegraphics[width=0.8\textwidth]{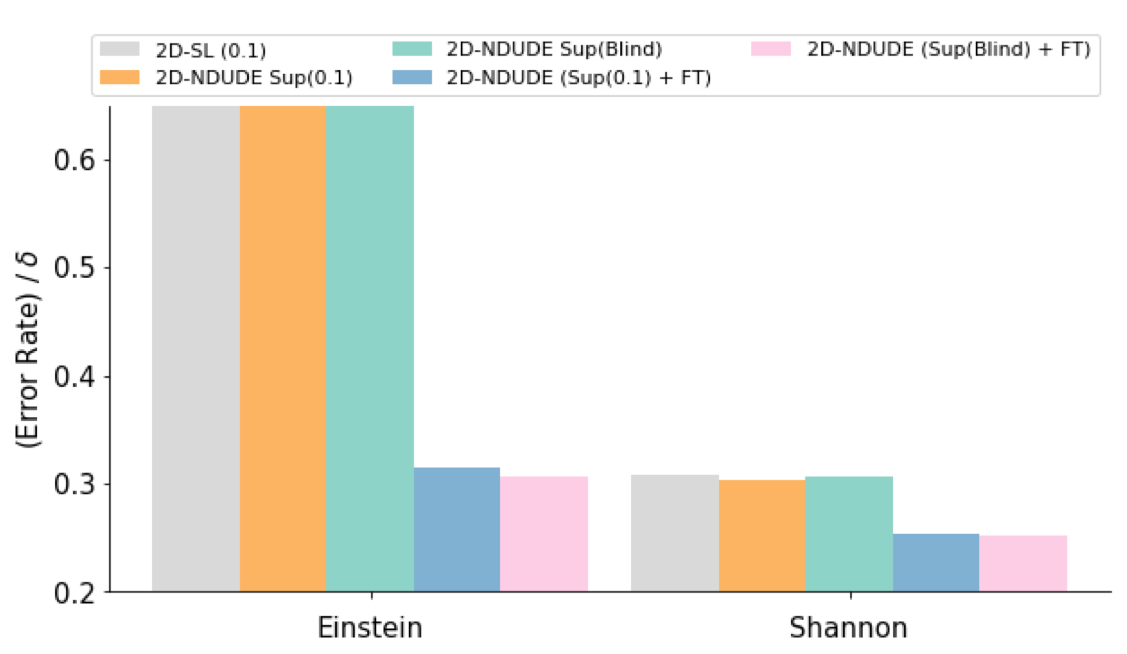}
   \caption{BER results on Set2 ($\delta=0.1$).}\label{fig:set2}
\end{figure*}

Figure \ref{fig:set2} shows the BER results for Set2 corrupted with true $\delta=0.1$ and stresses the effectiveness of the adaptive fine-tuning. Namely, since the images in Set2 have radically different characteristics compared to those in the training set, the BER differences between the supervised-only models and the fine-tuned models become larger. In other words, the fine-tuning step can successfully address the \textit{image mismatch} problem between the training data and the noisy images subject to denoising. 

\begin{figure*}[t!]
\centering
\includegraphics[width=0.8\textwidth]{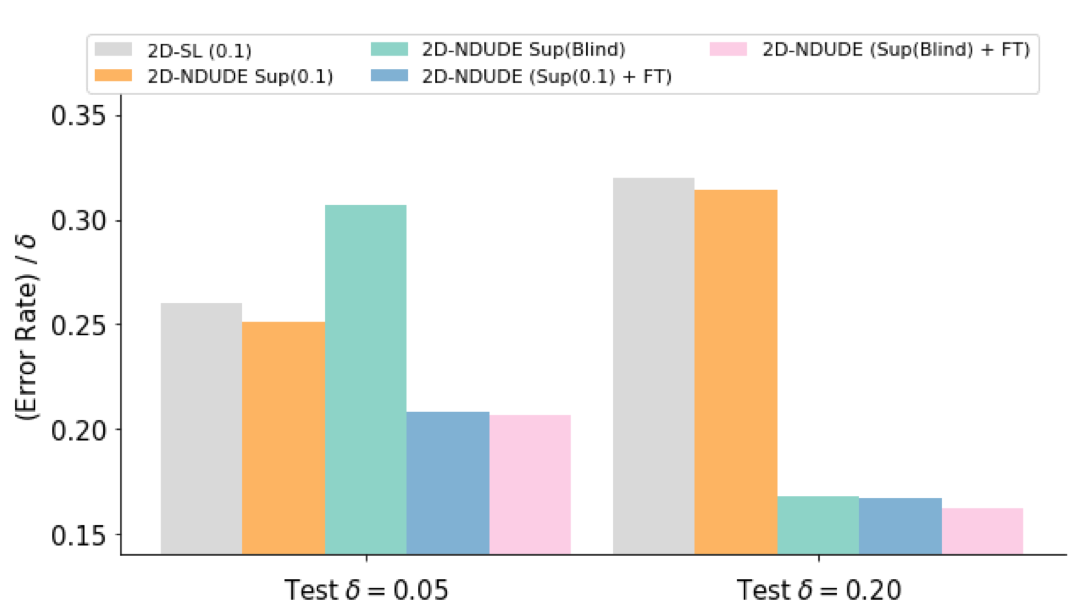}
          \caption{BER results on noise mismatch cases for BSD20.}
          \label{fig:bsd20_mismatch}
\end{figure*}

Figure \ref{fig:bsd20_mismatch} further shows the effectiveness of the adaptive fine-tuning for the \textit{noise mismatch} case. Namely, the supervised-only models, \texttt{2D-SL} and \texttt{2D-NDUDE(Sup)}, are trained on $\delta=0.1$, while the test noisy images in BSD20 are corrupted by BSC with either $\delta=0.05$ or $\delta=0.2$. We clearly observe that the noise mismatch makes the supervised-only models suffer from high errors, whereas the adaptive fine-tuning can significantly correct such mismatches. Note the blind training of \texttt{2D-NDUDE(Sup(Blind))} can help as long as the true $\delta$ is included in the composite noise range used for supervised training, as in $\delta=0.2$ case in Figure \ref{fig:bsd20_mismatch}, but, if $\delta$ is out of or on the boundary of the range, the BER significantly deteriorates as in $\delta=0.05$ case. 
% when the $\delta$ is included in the composite noise range used for training, as in $\delta=0.2$ case, but, if $\delta$ is out of or on the boundary of the range, the BER significantly deteriorates as in $\delta=0.05$ case.
% the experimental result of another mismatch case, the mismatch between model $\delta$ and test image $\delta$. In this experiment, all supervised models are trained on $\delta$=0.1 except for \texttt{2D-NDUDE (Sup(Blind))}, and we use the true $\delta$ of a test image for the adaptive fine tuning. For testing, we generated BSD20 on $\delta=\{0.05, 0.2\}$ . From Figure \ref{fig:bsd20_mismatch}, we obviously see that the adaptive fine tuning improves the denoising result even if the mismatch between model $\delta$ and test image $\delta$ exists. The results of supervised model have higher relative BERs than the results of the adaptive fine-tuning except for  \texttt{2D-NDUDE (Sup(Blind))}, which shows the competitive result on test $\delta=0.20$, but suffers on test $\delta$=0.05.

\begin{figure*}[h!]
\centering
\includegraphics[width=0.8\textwidth]{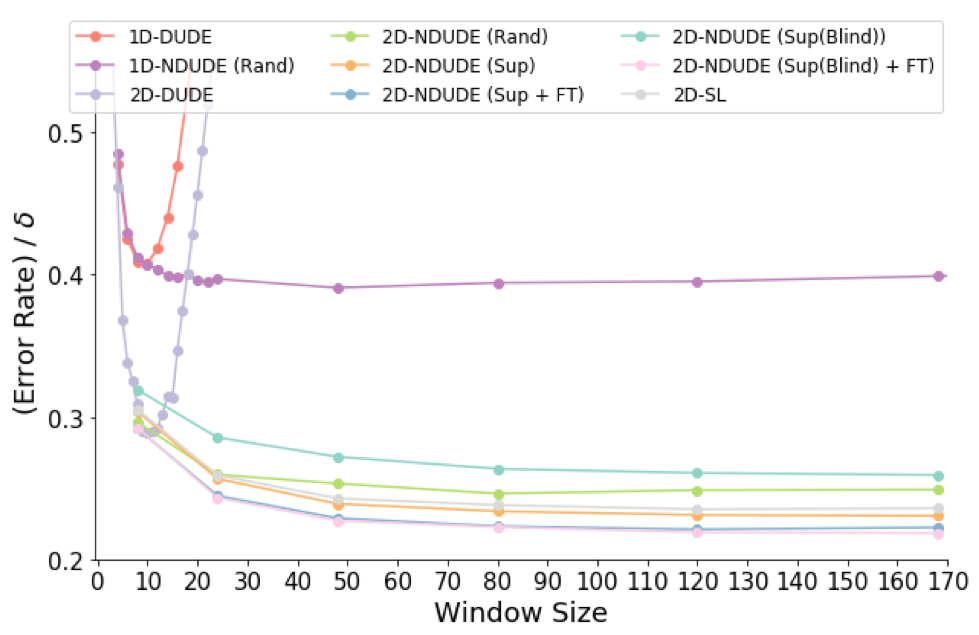}
          \caption{BER results on Set11 with varying $k$ ($\delta=0.1$).}
          \label{fig:set11_k}
\end{figure*}

Figure \ref{fig:set11_k} shows the BER results on Set11 with varying window size $k$ for $\delta=0.1$. As shown in \cite{MooMinLeeYoo16}, we note that both original DUDE methods are highly sensitive to $k$ values, while all the neural network based methods (also including \texttt{2D-SL}) become extremely robust to sufficiently large $k$ values. 

\begin{figure*}[h]
\captionsetup[subfigure]{justification=centering}
    \centering
      \begin{subfigure}{0.30\textwidth}
        \includegraphics[width=\textwidth]{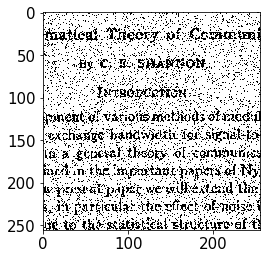}
          \caption{Noisy image \\ ($\delta$ = 0.10)}
      \end{subfigure}
      \begin{subfigure}{0.30\textwidth}
        \includegraphics[width=\textwidth]{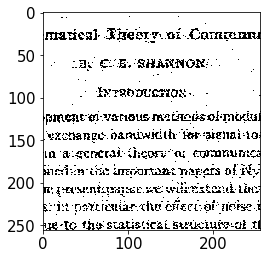}
          \caption{1D-NDUDE \\ (BER = 0.407)}
      \end{subfigure}
      \begin{subfigure}{0.30\textwidth}
        \includegraphics[width=\textwidth]{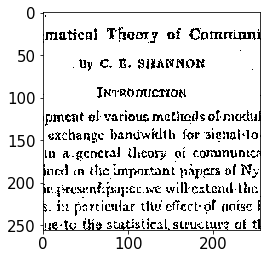}
          \caption{2D-NDUDE (Rand) \\ (BER = 0.267)}
      \end{subfigure}
      \begin{subfigure}{0.30\textwidth}
        \includegraphics[width=\textwidth]{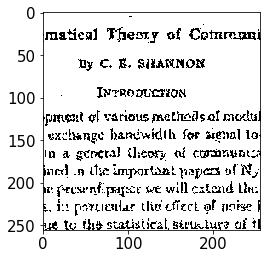}
          \caption{2D-NDUDE (Sup(Blind)) \\ (BER = 0.306)}
      \end{subfigure}
      \begin{subfigure}{0.30\textwidth}
        \includegraphics[width=\textwidth]{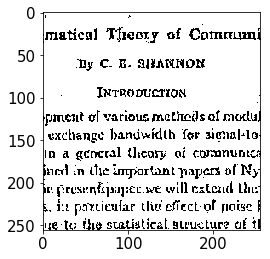}
          \caption{\textbf{2D-NDUDE (Sup(Blind)+FT)\\ (BER = 0.252)}}
      \end{subfigure}
      \begin{subfigure}{0.30\textwidth}
        \includegraphics[width=\textwidth]{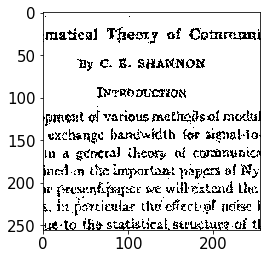}
          \caption{SL \\ (BER = 0.308)}
      \end{subfigure}
\caption{Visualization of denoising results of Shannon image in Set2 ($\delta=0.1$).}
\label{fig:shannon}
\end{figure*}

 Finally, Figure \ref{fig:shannon} visualizes and compares the denoising results on the Shannon image (a scanned text image) in Set2 for $\delta=0.1$. We note the denoising result of \texttt{2D-NDUDE(Sup(Blind)+FT)} achieves the lowest BER among all and results in the most readable denoised text. 
 
\subsection{DNA sequence denoising}
\subsubsection{Datasets and training details}

\begin{figure}[htb]
  \centering
  \centerline{\includegraphics[width=0.8\textwidth]{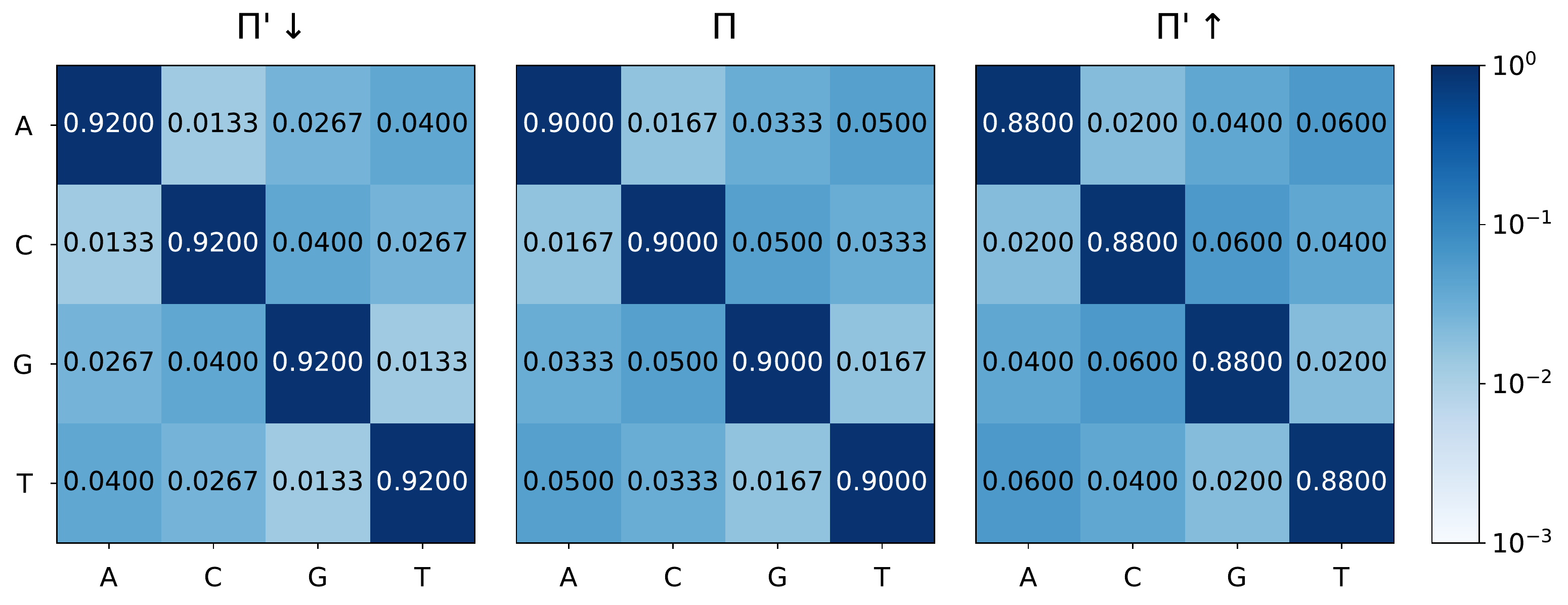}}
\vspace{-.3cm}
\caption{$\Pib$ for DNA sequence denoising.}
\label{fig:res_DNA_PI}
\end{figure}

We now carry out the experiment on the DNA sequence denoising. For our experiments, we used simulated next generation sequencing (NGS) DNA datasets, which were generated as follows. First, we obtained reference sequences $R$ from four organisms (Anaerocellum thermophilum Z-1320 DSM 6725 (AT), Bacteroides thetaiotaomicron VPI-5482 (BT), Bacteroides vulgatus ATCC 8482 (BV), Caldicellulosiruptor saccharolyticus DSM 8903 (CS)), targeting V3 hypervariable regions. Second, noiseless NGS reads $x^n$ were randomly generated from each reference sequence in $R$, while the read lengths and the number of reads were set to 200 and 6000 according to \cite{LeeMooYooWei16}. Then, based on $\Pib$, a quaternary symmetric channel with $\delta=0.1$ (Figure~\ref{fig:res_DNA_PI} middle), we obtained the corresponding noisy sequence $Z^n$ by inducing substitution errors.

In contrast to the image denoising, representative reference sequences of different species are usually available for noisy NGS DNA datasets. Although the reference sequences may slightly differ with corresponding noisy sequences due to individual single nucleotide polymorphism differences, they provide a good approximation of the clean data. In view of this situation, for generating the supervised datasets $\mathcal{D}$, we first produced $\tilde{R}$, different versions of reference sequences of the four organisms (AT, BT, BV and CS), by randomly mutating the original reference sequences $R$ by 1\%. Next, we obtained sets of clean and noisy read pairs by following the same procedures above with $\tilde{R}$ in each epoch of supervised training. For the experiments with the mismatched noise models, we also generated supervised datasets using $\Pib'\downarrow$ and $\Pib'\uparrow$, which have relatively 20\% lower and higher error rates than the original $\Pib$ (Figure~\ref{fig:res_DNA_PI} left and right). In addition, for the experiments with the \emph{blind} supervised model, we randomly selected $\Pib'$, quaternary symmetric channel with $\delta$ in [0.05, 0.25] for each training iteration and used it for generating the supervised datasets.

All DNA sequence denoising experiments were done with PyTorch (version 0.4.1)\footnote{\texttt{https://pytorch.org}}. Our fully-connected neural network has 4 layers with 100 nodes in each hidden layer, and used ReLU activation function. Adam \cite{KinBa15} with learning rate of $10^{-3}$ was used for both supervised and pseudo-label training of random initialized model, and learning rate of $10^{-4}$ was used for adaptive fine-tuning. As in binary image denoising experiments, we tested 5 variants of Neural DUDE model, \ie, \texttt{NDUDE(Rand)}, \texttt{NDUDE(Sup)}, \texttt{NDUDE(Sup+FT)}, \texttt{NDUDE(Sup(Blind))}, and \texttt{NDUDE(Sup(Blind)+FT)}. To evaluate the denoising performance of the proposed model, we compared its performance with two other alternatives: \texttt{DUDE} and \texttt{SL}. As in the case of image denoising, \texttt{SL} represents a vanilla supervised model which directly learns a mapping $\mcZ^{2k+1}\rightarrow\mcXhat$; we used the same fully-connected neural network architecture, optimizer, and learning rate for \texttt{SL} as the Neural DUDE models. Note all models are 1D models in the DNA experiments. 

\begin{figure*}[h!]

    % &   \multirow{2}{\hsize}[32mm]{
        \includegraphics[width=0.7\textwidth]{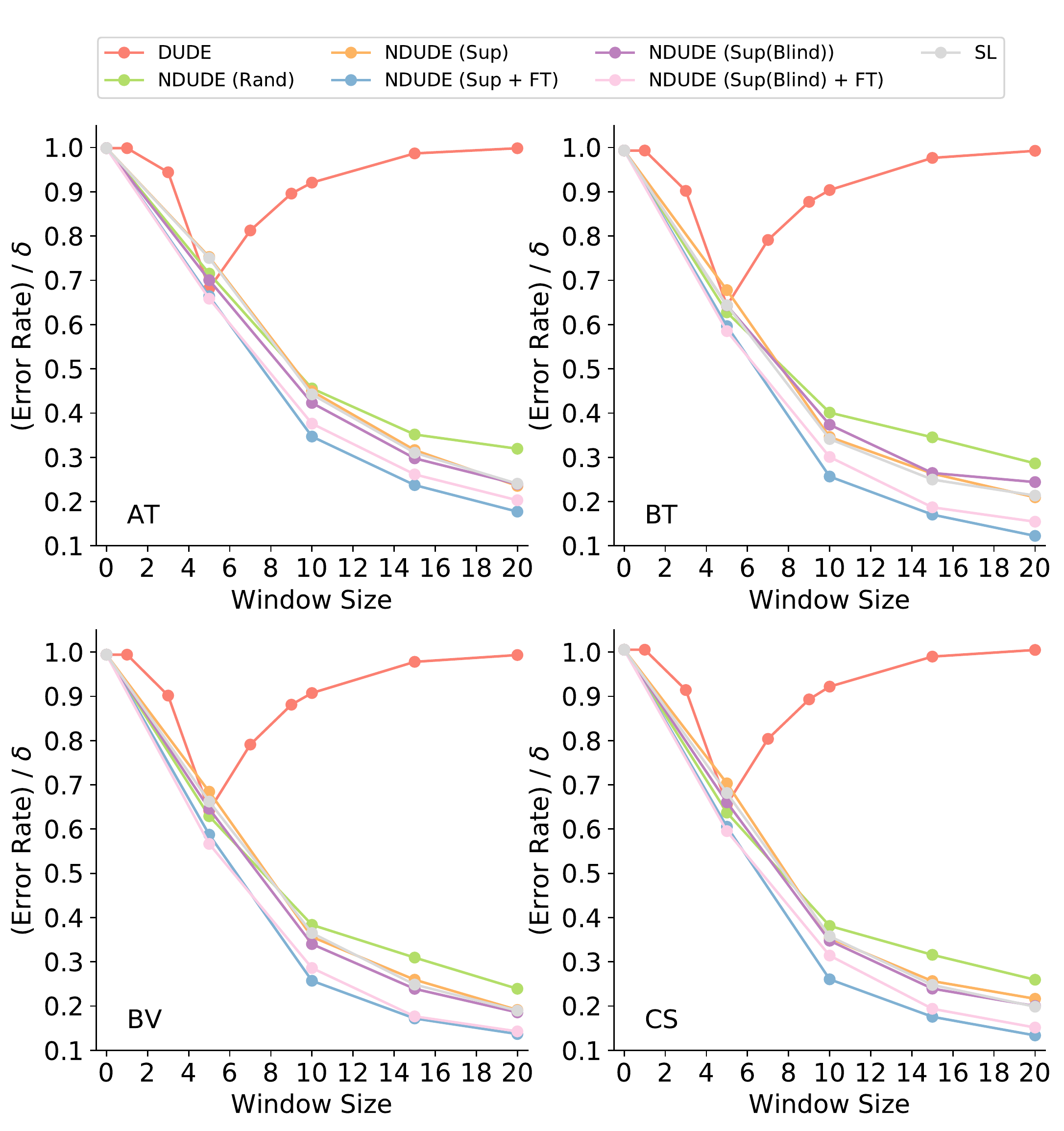}
        \centering
         \caption{Results on various window size $k$.}
        \label{fig:res_DNA1} 
% \caption{Experimental results on DNA seqeunce denoising.}
\label{fig:process1}
\end{figure*}

% \begin{figure}[htb]
%   \centering
%   \centerline{\includegraphics[width=9.0cm]{figures/AAAI2018_DNA_Result1.pdf}}
% \vspace{-.3cm}
% \caption{DNA sequence denoising results.}
% \label{fig:res_DNA1}
% \end{figure}

\subsubsection{Results}

% \begin{figure*}[h]
%     \begin{tabular}{p{0.48\textwidth}p{0.48\textwidth}}
%     \centering
%     \includegraphics[height=3.5cm]{figures/AAAI2018_DNA_PI.pdf}
%     \subcaption{$\Pib$ for DNA sequence denoising.}
%         \label{fig:res_DNA_PI}

%     \includegraphics[height=5.cm]{figures/AAAI2018_DNA_Result2.pdf}
%     \centering
%     \subcaption{Results on mismatched and blind noise models.}
%     \label{fig:res_DNA2}
    
%     % &   \multirow{2}{\hsize}[32mm]{
%         \includegraphics[height=8.8cm]{figures/AAAI2018_DNA_Result1.pdf}
%     %     \centering
%          \subcaption{Results on various window size $k$.}
%         \label{fig:res_DNA1} 
%         %                     }
%         % \\[-4mm]
    
%         % &    
%         \end{tabular}
% \caption{Experimental results on DNA seqeunce denoising.}
% \label{fig:process1}
% \end{figure*}

Figure~\ref{fig:res_DNA1} shows the denoising results with respect to the sliding window size $k$. The vertical axis is the error rate normalized with $\delta$ after denoising. For all the datasets from four organisms, we could clearly observe the effectiveness of the proposed methodology. First, in contrast to \texttt{DUDE}, \texttt{NDUDE(Rand)} again achieves significantly better performance by sharing the information from similar contexts as $k$ becomes larger. Note this is consistent with the results of \cite{MooMinLeeYoo16} for DNA sequence denoising. Second, the supervised training with clean and noisy read pairs from $\tilde{R}$ improved the denoising performances, and both \texttt{NDUDE(Sup)} and \texttt{SL} got much better than \texttt{NDUDE(Rand)}. Furthermore, combined with adaptive fine-tuning, \texttt{NDUDE(Sup+FT)} additionally provided relative 25.8\% improvements on average over \texttt{NDUDE(Sup)} or \texttt{SL} in the four organisms and $k$ in $\{5, 10, 15, 20\}$. Both \texttt{NDUDE(Sup(Blind))} and \texttt{NDUDE(Sup(Blind)+FT)} also achieved results comparable to \texttt{NDUDE(Sup)} and \texttt{NDUDE(Sup+FT)}, respectively. 

% \begin{figure}[htb]
%   \centering
%   \centerline{\includegraphics[width=9.0cm]{figures/AAAI2018_DNA_Result2.pdf}}
% \vspace{-.3cm}
% \caption{$\Pib$ for DNA sequence denoising.}
% \label{fig:res_DNA2}
% \end{figure}

% We now compare the robustness of \texttt{NDUDE(Sup+FT)} to the mismatched and even blind noise models in DNA sequence denoising. 

\begin{figure*}[h]
    \includegraphics[width=0.8\textwidth]{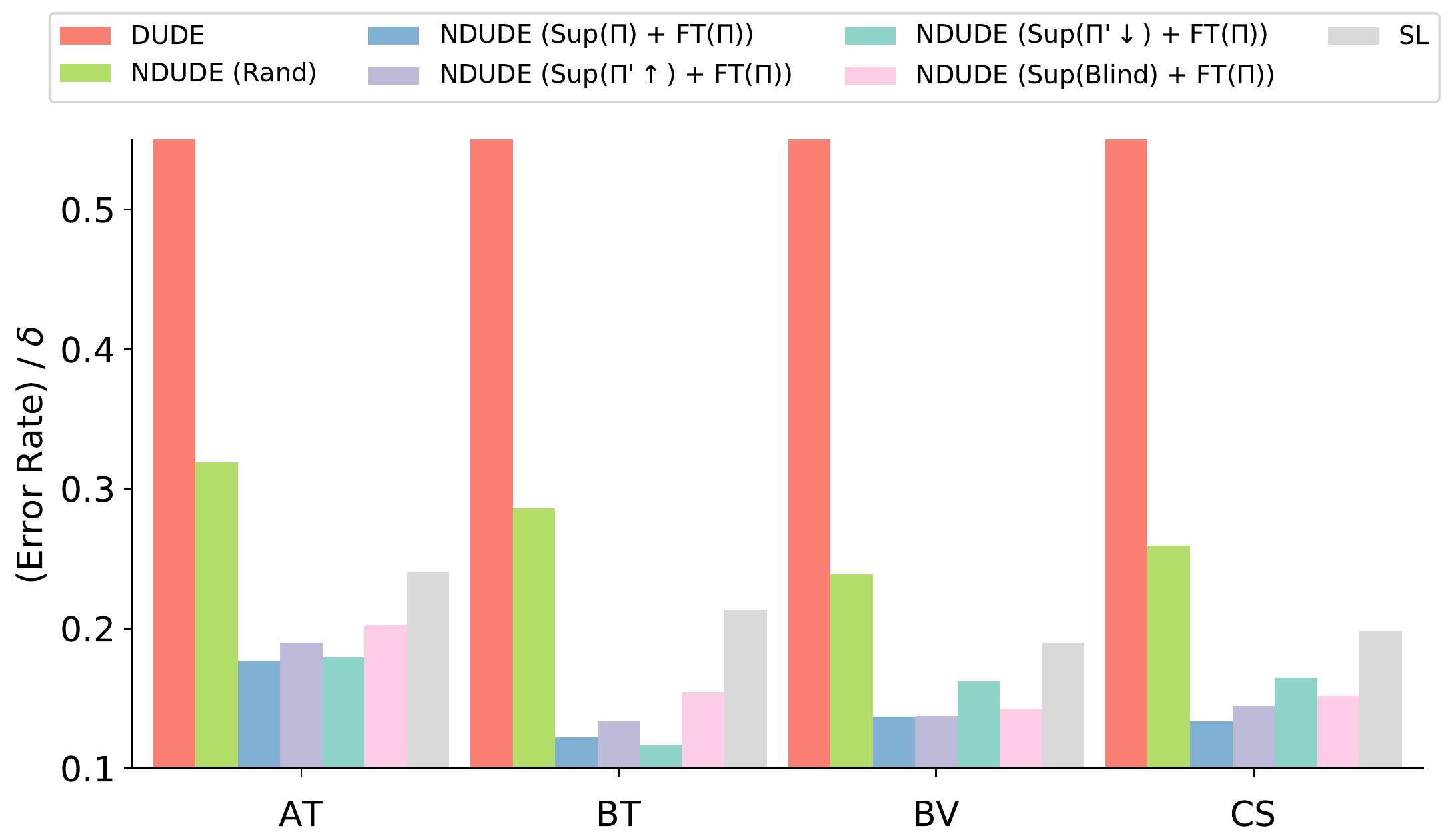}
    \centering
    \caption{Results on mismatched and blind noise models.}
    \label{fig:res_DNA2}
\end{figure*}

Figure~\ref{fig:res_DNA2} shows the overall denoising results for each organism dataset and for the best $k$ values for each scheme, \ie, $k=5$ for \texttt{DUDE} and $k=20$ for all the Neural DUDE variants. \texttt{FT($\Pib$)} stands for the fine-tuning done with the correct channel $\Pib$. Again, we could see that when the correct $\Pib$ is used for adaptive fine-tuning, the mismatched (with \texttt{Sup($\Pib'\downarrow$)} or \texttt{Sup($\Pib'\uparrow$)} notations) or blindly trained (with \texttt{Sup(Blind)} notation) models can still achieve relative 28.5\% improvements on average compared to the performance of \texttt{NDUDE(Rand)}.
% supervised training with the mismatched and blind noise models still provides relative 28.5\% improvements on average compared to the performance of \texttt{NDUDE(Rand)}. 
Moreover, \texttt{NDUDE(Sup(Blind)+FT)}, the blind supervised pre-training combined with the adaptive fine-tuning is also robust to the context size and consistently outperforms \texttt{NDUDE(Rand)} for all $k$ in $\{5, 10, 15, 20\}$, hence, confirms the strong adaptivity of our method. 

\section{Conclusions} 
We improved the original Neural DUDE by making the supervised learning framework compatible with the adaptive fine-tuning. In results, we achieved much better denoising performances than those of \cite{MooMinLeeYoo16} as well as the robustness with respect to the mismatch of the image characteristics or the noise levels, which are prevalent in practice. Such robustness makes the algorithm practical as maintaining a single blindly trained supervised model would suffice to achieve good denoising performance, provided that the correct $\Pib$ can be used during fine-tuning.

{\small
\bibliographystyle{plain}
\bibliography{bibfile}
}
\end{document}